\documentclass[conference]{IEEEtran}
\IEEEoverridecommandlockouts
\usepackage{cite}
\usepackage{amsmath,amssymb,amsfonts}
\usepackage{graphicx}
\usepackage{booktabs}
\usepackage{textcomp}
\usepackage{xcolor}
\usepackage{url}
\def\BibTeX{{\rm B\kern-.05em{\sc i\kern-.025em b}\kern-.08em
    T\kern-.1667em\lower.7ex\hbox{E}\kern-.125emX}}

\definecolor{linkteal}{rgb}{0.15,0.45,0.55}
\usepackage[colorlinks=true,
            urlcolor=linkteal,
            linkcolor=black,
            citecolor=black]{hyperref}
            
\begin{document}

\title{How Should a Simulation-to-Reality Transfer Budget Be Spent?\\
\thanks{$^{*}$These authors contributed equally and share first authorship. Submitted to the First Workshop on Sim2Real and Classical Control: From Rigorous Theory to Data-Driven Robotics, IROS 2026.}
}

\author{\IEEEauthorblockN{Syed Hamzah Rizvi$^{*}$}
\IEEEauthorblockA{\textit{Purdue University} \\
West Lafayette, USA \\
syed105@purdue.edu}
\and
\IEEEauthorblockN{Yash Vardhan Tomar$^{*}$}
\IEEEauthorblockA{\textit{Purdue University} \\
West Lafayette, USA \\
tomar4@purdue.edu}
}

\maketitle

\begin{abstract}
Simulation-to-reality transfer, often called sim-to-real transfer, is a central challenge in robot learning. Yet, the tradeoff between measuring a system more accurately and training over a broader range of simulated dynamics is still poorly understood. In this work, we focused on the allocation of real-robot measurement time between system identification and domain randomization. We studied this tradeoff in a controlled sim-to-sim pendulum setting, where a hidden-parameter model stands in for the physical robot, and the experiment sweeps identification rollouts against the width of the randomization distribution. Across the reality gaps and noise levels we tested, the measurement budget did most of the work. A small number of identification rollouts closed most of the transfer gap, and once any real data was available, policies performed best when trained at the estimated parameters rather than over a widened randomization band. Broad randomization that contained the true system still did not substitute for measurement. These results hold in a benign regime where the dynamics are identifiable and only two parameters are unknown, so structural model mismatch remains the setting where randomization breadth may become more valuable. Overall, our results suggest that sim-to-real pipelines should first measure the parameters they can and reserve randomization for the uncertainty that remains.
\end{abstract}


\section{Introduction}
Reinforcement learning (RL) offers a natural framework for training control policies through repeated interaction. In robotics, however, those interactions are often too expensive, slow, or unsafe to collect directly on hardware. Simulation provides an attractive alternative: policies can be trained at scale before deployment on the physical system. This introduces a central difficulty: transfer. A policy that performs well in simulation may fail on the real robot because the simulator only approximates the true dynamics, and the learned policy can exploit modeling errors that do not exist on hardware. This mismatch is known as the \emph{reality gap}~\cite{tobin2017}, and reducing it remains a central challenge in sim-to-real robot learning.

Two standard approaches address this gap. System identification uses measurements from the real platform to estimate physical parameters and tune the simulator toward the target dynamics~\cite{chebotar2019}. Domain randomization takes the opposite approach: rather than committing to a single parameter estimate, it trains policies over a distribution of randomized simulators, with the goal of learning a behavior that remains robust when deployed on the real system~\cite{tobin2017,peng2018}. Both approaches are widely used, but they rely on the same scarce resource: interaction with the physical robot.

This shared cost makes the choice between system identification and domain randomization a budget-allocation problem. Each real rollout can be used to estimate the target system more accurately, or it can justify a broader randomization distribution for policy training. When only a small number of hardware rollouts are available, a practitioner must decide how much of that budget to spend on measurement and how much uncertainty to leave to randomization. Existing work typically studies these methods separately, leaving this allocation question underexplored. Truong \emph{et al.}~\cite{truong2021} show that higher-fidelity simulation is not always worth its cost, suggesting that the value of measurement should be evaluated directly rather than assumed. This motivates our central question: \emph{given a fixed measurement budget, how should that budget be divided between system identification and domain randomization?}

We answer this question through direct measurement in a controlled sim-to-sim study. A second simulator, parameterized by hidden dynamics, serves as the target system. This design lets us run the repeated trials needed for a careful comparison while keeping the target dynamics unknown to the learner. We vary two quantities independently: $n$, the number of target-system rollouts allocated to identification, and $w$, the width of the randomization distribution. We then evaluate the resulting surface of zero-shot return, defined as policy performance on the hidden target system without further tuning.

We do not propose a new algorithm. Instead, our contribution is to make the identification-versus-randomization tradeoff measurable under a fixed budget. First, we formulate the choice as a single budget-allocation problem and map the full return surface (Sec.~\ref{sec:results}). Second, using paired-bootstrap significance over ten seeds, we show that measurement dominates in this identifiable regime: a small number of identification rollouts closes most of the transfer gap, and once any target-system data is available, training at the recovered parameters significantly outperforms randomizing around them. Third, we test a natural alternative by training over a wide, truth-bracketing prior-range randomization distribution; even this broader distribution does not substitute for measurement. We further confirm the same ordering across additional reality gaps and noise levels (Table~\ref{tab:robust}).

We deliberately study a favorable setting in which the relevant dynamics are identifiable, the simulator family contains the target system, and only two physical parameters are unknown. We use this controlled regime as a baseline for the harder setting where structural model mismatch prevents any parameter estimate from making the simulator fully reproduce the target dynamics.

\subsection{Related work}
Domain randomization for sim-to-real transfer was popularized for perception by
Tobin \emph{et al.}~\cite{tobin2017} and for dynamics by Peng \emph{et
al.}~\cite{peng2018}; broad surveys are available~\cite{zhao2020,muratore2021review}.
A closely related line of work spends a real-world budget on the randomization
distribution itself, adapting it from data: SimOpt closes the sim-to-real loop by
updating the randomization from real rollouts~\cite{chebotar2019}; BayRn uses Bayesian optimization over the source
distribution from sparse real data~\cite{bayrn}; DROPO fits a randomization
distribution from a limited offline dataset~\cite{dropo}; auto-tuned transfer
iteratively shifts the simulator parameters toward the real
system~\cite{autotuned}; and Vuong \emph{et al.} optimize the randomization
parameters directly for real-world return~\cite{vuong}. Closest in spirit and
platform, Shakerimov \emph{et al.}~\cite{shakerimov} show that as few as 20--50
real episodes substantially improve a randomization-trained agent on a pendulum.
On the theory side, Chen \emph{et al.}~\cite{chen2021} give bounds on the
sim-to-real gap of randomization and the conditions under which it can succeed
without real samples.

Two empirical studies reach conclusions that appear to oppose ours and are worth
engaging directly. AdaptSim~\cite{adaptsim} argues that a task-driven simulator
adaptation can outperform exact dynamics matching when the sim-to-real gap is
irreducible, and Valassakis \emph{et al.}~\cite{valassakis} find that simple
random-force injection is competitive with online adaptation under the model
mismatch. We read the apparent tension as a difference in regime. Both works
operate where a structural mismatch makes the true system unpresentable by the
training model, whereas our testbed is well specified and identifiable, so our
finding appears to sit at the well-specified end of the spectrum they occupy. Relative to
all of this work, we hold the learner and task fixed, and sweep the identification
budget against the randomization width, so the return surface isolates how a fixed
measurement budget is best spent.

\section{Methodology}

\subsection{Setup}
We use two pendulum simulators. The reality simulator $R$ has fixed but hidden
physical parameters, the bob mass $m$ and the rod length $\ell$, which we collect
into a vector $\theta=(m,\ell)$; it is our proxy for the physical robot. We set its
true, hidden parameters to $\theta^\star=(m^\star,\ell^\star)=(2.0,1.5)$, well away
from the nominal prior $\theta_0=(1.0,1.0)$ that an engineer would assume before
any measurement, so the gap actually costs something. This offset was chosen
deliberately: a smaller gap left too little headroom for any method to separate,
so the specific numbers below are specific to this calibration. The training
simulator $T$
exposes those same parameters as adjustable. Gravity, timestep, torque limit,
and speed limit are shared and fixed across both simulators. Both simulators use
the same Pendulum dynamics; the only thing hidden from training is the value of
$\theta^\star$. This is a deliberate choice that we return to in
Section~\ref{sec:limitations}.

\subsection{Budget and Allocation}
A run is defined by a real-data budget $n$, the number of rollouts on $R$ spent
on identification, and a randomization width $w$. Identification collects $n$
rollouts on $R$ under a random-torque excitation policy (torques drawn i.i.d.\
from the uniform distribution $\mathcal{U}(-2,2)$\,N\,m, the full torque limit of
the platform) and fits the training parameters $\theta$ to $R$ by a grid search
over physically plausible bounds $[0.3,3.0]^2$ ($61\times61$, fine enough that grid
quantization stays well below the sensor noise) that minimizes one-step
angular-velocity prediction error. This
returns a point estimate $\hat\theta=(\hat m,\hat\ell)$, whose quality we summarize
by the identification error $\lVert\hat\theta-\theta^\star\rVert$. We add Gaussian sensor noise of standard
deviation $\sigma=1.0$ to the measured angular velocity, so the identification
error falls roughly as $1/\sqrt{n}$ instead of collapsing after a handful of
rollouts. We emphasize that $n$ counts rollouts of
this random excitation policy; informative or Fisher-optimal
excitation~\cite{spiactive} would identify the parameters from fewer rollouts, so
our budget axis is a conservative, excitation-dependent estimate of how fast
identification closes the gap. The $n$ rollouts also cost real-system time and
carry deployment risk (a random policy on hardware), beyond their mere count.
Setting $n=0$ skips identification and trains around the prior. The width $w$
sets the spread of the randomization distribution as a fraction of the point
estimate: each episode resamples $(m,\ell)$ component-wise and uniformly from
$[\hat\theta(1-w),\,\hat\theta(1+w)]$ (clipped to the same $[0.3,3.0]$ bounds),
so $w=0$ trains at the estimate alone and larger $w$ trains over a wider band
centered on the estimate. We cap $w$ at $0.5$; wider settings drove
parameters against the clip bounds and produced pathological dynamics. However, we note that
because the band is relative to $\hat\theta$, at $n=0$ it is centered on the prior
$\theta_0$, so even $w=0.5$ samples mass only up to $1.5$ and never reaches the
true value $m^\star=2.0$. We return to this in Section~\ref{sec:limitations}.
Randomization itself runs entirely in simulation and costs no real-robot time.

\subsection{Training and Evaluation}
For each $(n,w)$ cell, we estimate the parameters, train a soft actor-critic
(SAC) policy in the randomized $T$ for $75,000$ steps (comfortably enough for SAC
to converge on the nominal task), and evaluate zero-shot return on the hidden $R$
over 20 deterministic episodes at the standard 200-step horizon (the return is the
pendulum quadratic cost summed over the episode, so values are negative and larger
is better; each seed uses its own fixed set of evaluation initial conditions). We use SAC because
it is sample-efficient and reliably solves the nominal pendulum; hyperparameters
follow standard values (learning rate $10^{-3}$, batch size 256, discount
$0.99$, replay buffer $10^5$, target smoothing $\tau=0.005$). We sweep $n$ from no
data to a substantial budget, $n\in\{0,5,10,25,50\}$, and $w$ from the point
estimate to a wide band, $w\in\{0,0.1,0.2,0.3,0.5\}$, over ten seeds each (enough
to resolve the noisy low-budget rows), for 250 runs in total. For the
load-bearing comparisons below, we report paired contrasts across the ten matched
seeds with $95\%$ confidence intervals from a paired bootstrap
($2{\times}10^4$ resamples). We define \emph{significant} as an interval excluding zero.

\section{Experiments and Results}
\label{sec:results}

\begin{figure}[h]
\centerline{\includegraphics[width=\columnwidth]{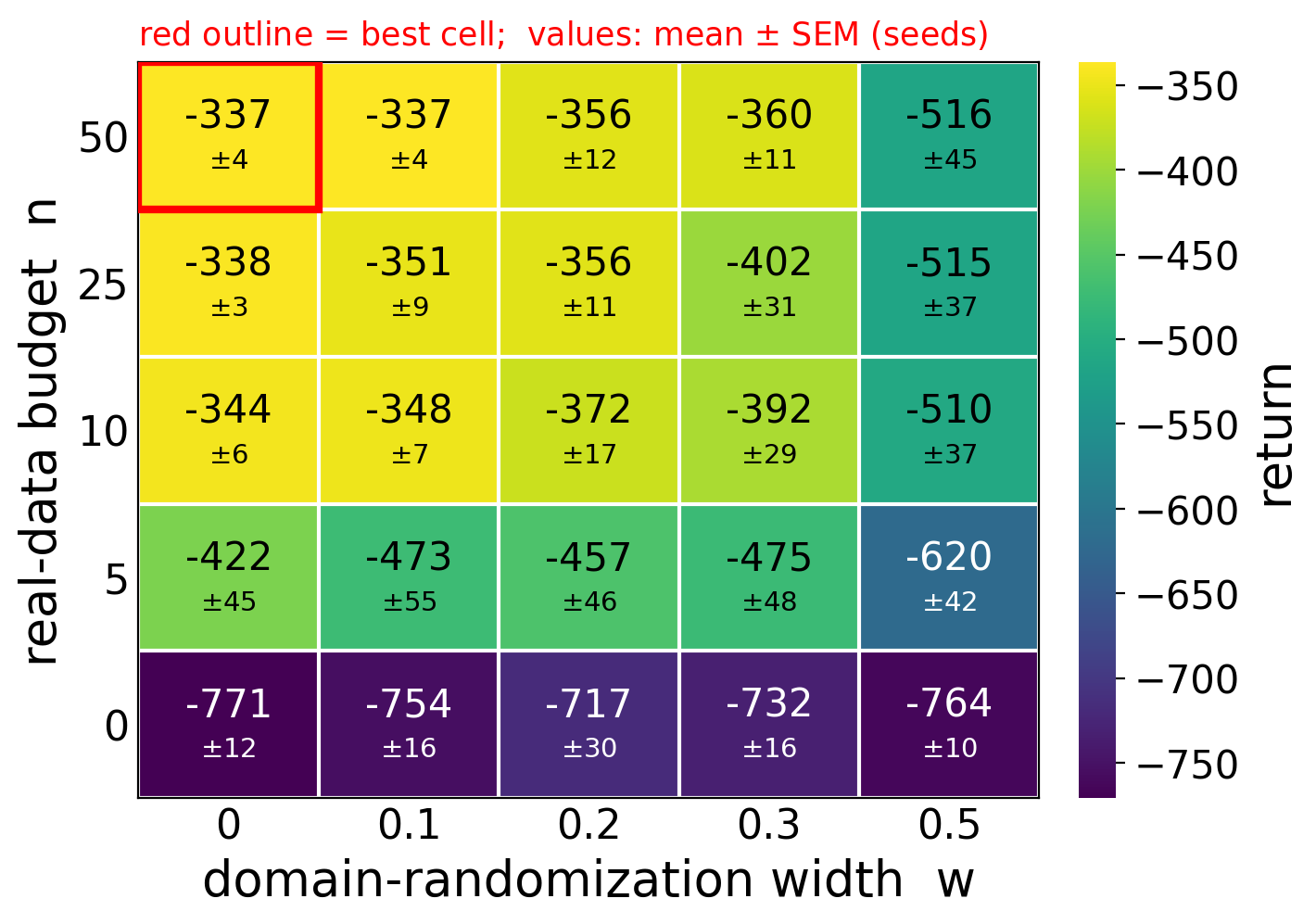}}
\caption{Mean zero-shot return on the hidden system $R$ over the grid of
real-data budget $n$ and randomization width $w$. Cell values are mean $\pm$ SEM
over ten seeds (these SEM values, not the paired-bootstrap CIs used for the
significance statements in the text, are what the annotations show). The best cell
($n=50$, $w=0$) is outlined in red. Return improves sharply with the first few
identification rollouts; added width does not help and significantly lowers return
at $w=0.5$ for every positive budget.}
\label{fig:heatmap}
\end{figure}

Fig.~\ref{fig:heatmap} shows the mean zero-shot return $\pm$ the standard error of the
mean, SEM, over ten seeds
over the whole grid. The structure is almost entirely vertical. The $n=0$ row
sits between $-717$ and $-771$ regardless of width, while every cell from $n=10$
upward sits near $-340$. The first identification rollouts do most of the work:
$n=5$ already lifts the $w=0$ column to about $-420$, and $n=10$ reaches $-344$;
the $n{=}0\!\to\!n{=}10$ jump at $w=0$ is $+427$ (95\%~CI $[+407,+448]$), by far
the largest effect in the study. At $n=0$, no identification occurs, and the
reported error of $1.12$ is simply the prior-to-truth distance
$\lVert\theta_0-\theta^\star\rVert$; with measurement, mean identification error
falls to $0.64$ at $n=50$, so the budget axis is doing real work and is not
saturated. The error settles above zero instead of vanishing. Two things keep it
there: the additive sensor noise, which also enters the one-step predictor's
inputs and biases the least-squares fit, and the finite identification grid.

Reading along any row, moving right (widening the randomization band around the
current estimate) never improves return. At every positive budget the
widest band $w=0.5$ is significantly worse than $w=0$ (paired
$\Delta\!\approx\!165$--$198$, all 95\%~CIs excluding zero); at $n=0$, by
contrast, $w=0.5$ and $w=0$ are statistically indistinguishable
($\Delta=-7$, 95\%~CI $[-34,+19]$), as the whole zero-budget row is poor.

\begin{figure}[t]
\centerline{\includegraphics[width=\columnwidth]{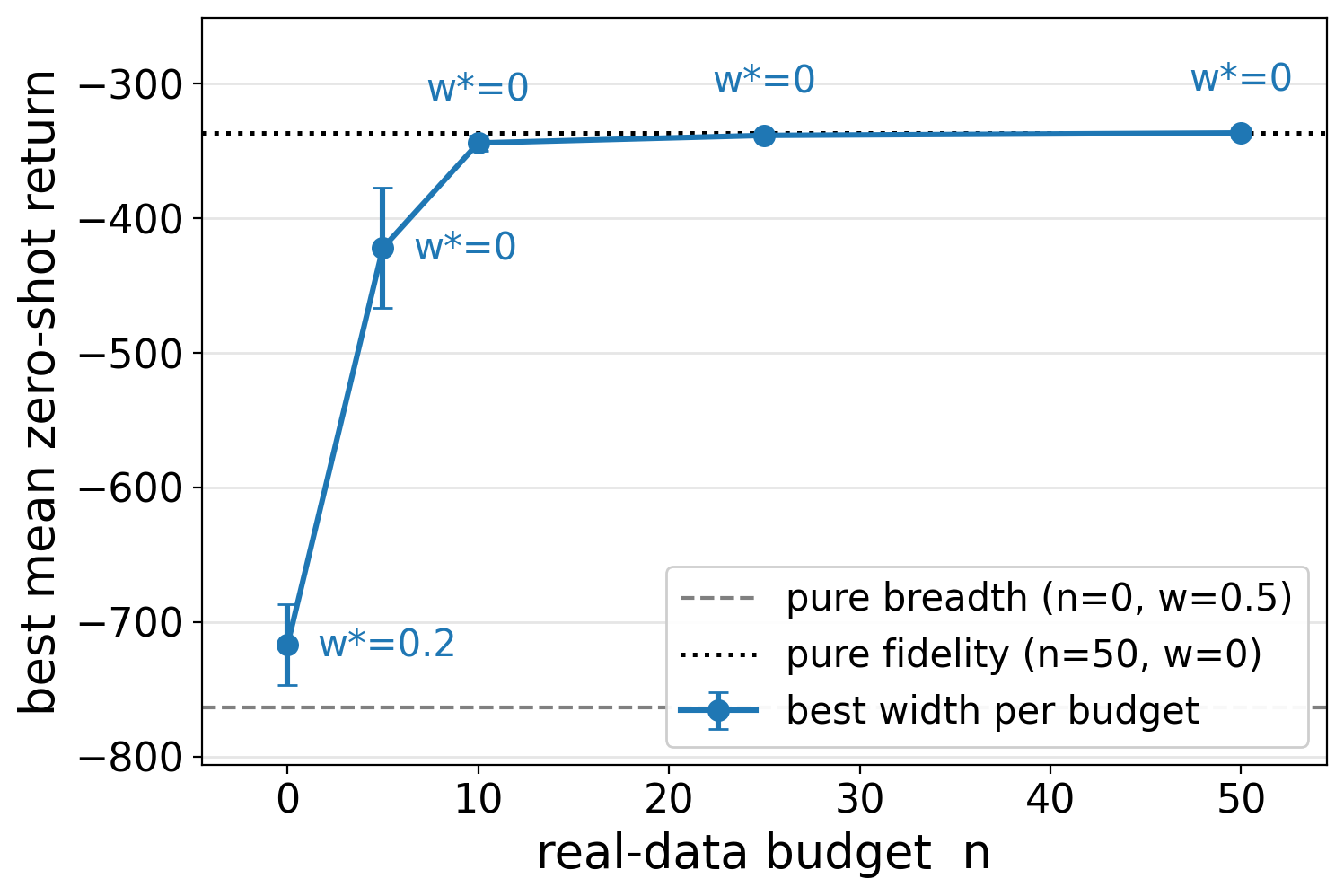}}
\caption{Best mean zero-shot return at each budget, with the chosen width
$w^\star$ labeled, against pure-breadth and pure-fidelity baselines. Error bars are
the standard error over ten seeds. Most of the gap closes by $n=10$, and the 
optimal width is zero once the budget is positive.}
\label{fig:pareto}
\end{figure}

Fig.~\ref{fig:pareto} takes the best width at each budget and plots it against
two references: pure breadth ($n=0$, $w=0.5$) and pure fidelity ($n=50$, $w=0$).
The pure-fidelity reference is, by construction, also the best cell in the grid.
The curve climbs steeply and then flattens; by $n=10$ it has reached the
pure-fidelity line at $-344$ (SEM~$6$) and stays there through $n=50$. The best
width is $w^\star=0.2$ only at $n=0$, but this gain is not significant
($\Delta=+54$ vs.\ $w=0$, 95\%~CI $[-4,+114]$): the zero-budget width effect is
within seed-to-seed variation. For every budget from $n=5$ on the best width is
$w^\star=0$, and the advantage of $w=0$ over even a narrow $w=0.2$ band is
significant ($\Delta\!\approx\!18$--$35$, all 95\%~CIs excluding zero). Once any
real data is in hand, the agent is best trained at the point estimate.

\subsection{A prior-range baseline and robustness checks}
Two follow-ups probe whether the ordering is an artifact of the
estimate-centered band or of the single operating point.

We add the standard wide-prior baseline the comparison most needs: at $n=0$, randomize $(m,\ell)$ uniformly over the full
bounds $[0.3,3.0]^2$, which do bracket the truth, independent of any
estimate. Over ten seeds, it returns $-749\pm18$, which is statistically
indistinguishable from estimate-centered breadth ($-764$; paired $\Delta=+15$,
95\%~CI $[-17,+46]$) and from the no-data point ($-771$), and it sits $413$ below
pure fidelity ($-337$; CI $[-442,-385]$). So the breadth's failure to substitute for measurement is not
a mechanical consequence of centering the band on a biased estimate. A
distribution that covers the true parameters still transfers poorly here, because
one policy trained to handle the whole plausible range is mediocre on any specific
system.

\subsection{Impact of Reality Gap and Sensor Noise} We repeat a reduced sweep at two further reality gaps and
two further sensor-noise levels (Table~\ref{tab:robust}). The qualitative ordering
holds at every operating point we tried. A positive budget significantly raises
return, and the widest band significantly lowers it at the full budget. Two
quantitative dependencies show up, both of the kind we would expect. The plateau in
the budget axis tracks the measurement noise. Identification reaches its floor by
about five rollouts at $\sigma=0.5$, but needs closer to fifty at $\sigma=2.0$. The $\approx\!10$-rollout knee is specific to the noise level and excitation, and shifts when either
changes. The small zero-budget benefit of breadth grows with the gap. It is
insignificant
at the main gap, significantly unhelpful at the small gap (where the prior
is already close), and significantly helpful at the large gap (where only a wide
band reaches toward the truth).

\par\vspace{8pt}
\noindent\begin{minipage}{\columnwidth}
\centering
\refstepcounter{table}\label{tab:robust}
{\small
\setlength{\tabcolsep}{0pt}
\renewcommand{\arraystretch}{1.08}
\begin{tabular*}{\columnwidth}{@{\extracolsep{\fill}}lcccc@{}}
\toprule
Operating point & Best & $\Delta$bud & $\Delta$wide & $w^\star_{0}$ \\
\midrule
Main $(2.0,1.5)$, $\sigma{=}1.0$ & $-337$ & \textbf{+434} & \textbf{+179} & $0$ \\
Small gap $(1.5,1.2)$ & $-226$ & \textbf{+29} & \textbf{+39} & $0$ \\
Large gap $(2.5,2.0)$ & $-573$ & \textbf{+377} & \textbf{+235} & \textbf{$>0$} \\
Low noise $\sigma{=}0.5$ & $-341$ & \textbf{+429} & \textbf{+223} & $0$ \\
High noise $\sigma{=}2.0$ & $-357$ & \textbf{+413} & \textbf{+106} & $0$ \\
\bottomrule
\end{tabular*}}
\par\vspace{6pt}
\noindent\parbox{\columnwidth}{\raggedright
\textbf{Table \thetable.} Robustness of the ordering across operating points.
Across every gap and noise level, a positive budget significantly raises return
and the widest band significantly lowers it. Bold values have 95\% CIs excluding
zero; reduced sweeps use 5 seeds.
}
\end{minipage}
\par\vspace{8pt}

\section{Limitations}
\label{sec:limitations}

\subsection{The identification model is well specified.} Our identifier fits the same
dynamics equations that generate the reality simulator, so the estimator's model
class contains the true system exactly, and the only discrepancy is zero-mean
sensor noise that averages out with more rollouts. Under these conditions, 
identification is essentially guaranteed to converge, and there is no
structural model mismatch for randomization to absorb. In real sim-to-real,
the dominant error is precisely such mismatch (unmodeled friction, backlash,
latency, contact) that no amount of mass-length fitting can remove, and that is
the regime in which breadth is expected to earn its keep. Our finding supports
measure first for identifiable parameters, and it is not evidence that
fidelity beats breadth under model mismatch.

\subsection{Randomization is centered on the estimate.} Because the randomization band is relative to $\hat{\theta}$, it centers on the prior $\theta_{0}$ at $n=0$. Even at $w=0.5$, the mass is drawn entirely from $[0.5, 1.5]$. Consequently, the true value of $2.0$ receives zero probability, leaving the true system completely outside the training support (where the upper edge sits exactly at $1.5$). Viewed in isolation, this phenomenon can mistakenly look like a mechanical inability to cover reality. Every randomization claim here should be read as concerning randomization around the current point estimate, not broad
randomization over a plausible range. We do test that standard
prior-range baseline (Table~\ref{tab:robust} and the prior-range result in
Section III): it brackets the truth yet still does not substitute for measurement, so the
the estimate-centering choice is not what drives the result. A method that
adapts the band from data, instead of fixing it, remains yet to be tested here.

\subsection{The budget axis depends on excitation and calibration.} Identification uses
a random-torque policy, which excites the dynamics unevenly; informative
excitation~\cite{spiactive} would close the gap with fewer rollouts, so the
$\approx\!10$-rollout knee is a conservative, excitation-specific figure. Likewise
the reality gap, noise level, and $1/\sqrt{n}$ rate are each measured at one
operating point chosen to make the trade-off visible, so the numeric knees are
illustrative for this calibration only.

\subsection{Other notable caveats.} At large $w$, the band is clipped to physical bounds and
enters low-control-authority regimes, so wide randomization hurts partly
on extreme, clipped dynamics; low return at the widest $w$ may
also partly reflect under-training, since a wide band changes the effective task
and may need more than $75\,000$ steps. Our significance statements use paired
bootstrap intervals over seed means rather than per-episode tests. We vary the reality
gap and the noise level (Table~\ref{tab:robust}), but the study stays on a single
task and a single two-parameter dynamics family. Further, the best return
($\approx-340$) is short of an optimally controlled pendulum.

\section{Conclusion}
Simulation-to-reality transfer is difficult in part because two reasonable strategies compete for the same limited resource. One strategy uses real-robot data to estimate the target system more accurately. The other uses the same data budget to justify broader domain randomization. In this work, we treated these competing strategies as a budget-allocation problem. We studied a controlled, sim-to-sim pendulum setting in which the target dynamics were well specified, identifiable, and governed by two unknown parameters. In this setting, we found measurement mattered more than breadth. A small number of identification rollouts closed most of the transfer gap, after which additional rollouts gave diminishing returns. Once any real data was available, policies performed best when trained at the estimated parameters, while wider randomization bands consistently reduced return, even when they contained the true system. We therefore interpret the result narrowly: when the parameters that matter can be identified within the simulator, system identification should come before domain randomization. Future work should test where this ordering changes under structural model mismatch, thereby being under a setting where the simulator cannot represent the target system by parameter fitting alone.

\par\vspace{6pt}

\noindent\textbf{Data and code availability statement.} The depicted data and the used code can be found at \url{https://github.com/YTomar79/sim2real_budget}.


\begin{thebibliography}{00}
\bibitem{tobin2017} J. Tobin, R. Fong, A. Ray, J. Schneider, W. Zaremba, and P. Abbeel, ``Domain randomization for transferring deep neural networks from simulation to the real world,'' in \emph{Proc. IEEE/RSJ Int. Conf. Intell. Robots Syst. (IROS)}, 2017, pp. 23--30.
\bibitem{peng2018} X. B. Peng, M. Andrychowicz, W. Zaremba, and P. Abbeel, ``Sim-to-real transfer of robotic control with dynamics randomization,'' in \emph{Proc. IEEE Int. Conf. Robot. Autom. (ICRA)}, 2018, pp. 3803--3810.
\bibitem{chebotar2019} Y. Chebotar, A. Handa, V. Makoviychuk, M. Macklin, J. Issac, N. Ratliff, and D. Fox, ``Closing the sim-to-real loop: Adapting simulation randomization with real world experience,'' in \emph{Proc. IEEE Int. Conf. Robot. Autom. (ICRA)}, 2019, pp. 8973--8979.
\bibitem{truong2021} J. Truong, S. Chernova, and D. Batra, ``Rethinking sim2real: Lower fidelity simulation leads to higher sim2real transfer in navigation,'' in \emph{Proc. Conf. Robot Learn. (CoRL)}, 2021.
\bibitem{zhao2020} W. Zhao, J. P. Queralta, and T. Westerlund, ``Sim-to-real transfer in deep reinforcement learning for robotics: A survey,'' in \emph{Proc. IEEE Symp. Series Comput. Intell. (SSCI)}, 2020, pp. 737--744.
\bibitem{bayrn} F. Muratore, C. Eilers, M. Gienger, and J. Peters, ``Data-efficient domain randomization with Bayesian optimization,'' \emph{IEEE Robot. Autom. Lett.}, vol. 6, no. 2, pp. 911--918, 2021.
\bibitem{dropo} G. Tiboni, K. Arndt, and V. Kyrki, ``DROPO: Sim-to-real transfer with offline domain randomization,'' \emph{Robot. Auton. Syst.}, vol. 166, 2023.
\bibitem{autotuned} Y. Du, O. Watkins, T. Darrell, P. Abbeel, and D. Pathak, ``Auto-tuned sim-to-real transfer,'' in \emph{Proc. IEEE Int. Conf. Robot. Autom. (ICRA)}, 2021, pp. 1290--1296.
\bibitem{vuong} Q. Vuong, S. Vikram, H. Su, S. Gao, and H. I. Christensen, ``How to pick the domain randomization parameters for sim-to-real transfer of reinforcement learning policies?'' \emph{arXiv preprint arXiv:1903.11774}, 2019.
\bibitem{shakerimov} A. Shakerimov, T. Alizadeh, and H. A. Varol, ``Efficient sim-to-real transfer in reinforcement learning through domain randomization and domain adaptation,'' \emph{IEEE Access}, vol. 11, 2023.
\bibitem{chen2021} X. Chen, J. Hu, C. Jin, L. Li, and L. Wang, ``Understanding domain randomization for sim-to-real transfer,'' in \emph{Proc. Int. Conf. Learn. Represent. (ICLR)}, 2022.
\bibitem{adaptsim} A. Z. Ren, H. Dai, B. Burchfiel, and A. Majumdar, ``AdaptSim: Task-driven simulation adaptation for sim-to-real transfer,'' in \emph{Proc. Conf. Robot Learn. (CoRL)}, 2023.
\bibitem{valassakis} E. Valassakis, Z. Ding, and E. Johns, ``Crossing the gap: A deep dive into zero-shot sim-to-real transfer for dynamics,'' in \emph{Proc. IEEE/RSJ Int. Conf. Intell. Robots Syst. (IROS)}, 2020, pp. 5372--5379.
\bibitem{muratore2021review} F. Muratore, F. Ramos, G. Turk, W. Yu, M. Gienger, and J. Peters, ``Robot learning from randomized simulations: A review,'' \emph{Front. Robot. AI}, vol. 9, 2022.
\bibitem{spiactive} M. Sobanbabu, G. He, T. He, Y. Yang, and G. Shi, ``Sampling-based system identification with active exploration for legged robot sim2real learning,'' in \emph{Proc. Conf. Robot Learn. (CoRL)}, 2025. [Online]. Available: arXiv:2505.14266.
\end{thebibliography}
\end{document}